\documentclass{article}
\usepackage{epsfig}
\usepackage{graphicx}
\usepackage{amsmath}
\usepackage{amssymb}
\usepackage{subcaption}
\usepackage{textcomp}
\usepackage{gensymb}
\usepackage{mathtools}
\usepackage{amssymb}
\usepackage{wrapfig}
\usepackage{lipsum}
\usepackage{diagbox}
\usepackage{booktabs}
\usepackage{stfloats}
\usepackage{multirow}
\usepackage{tabularx}
\newcolumntype{Y}{>{\centering\arraybackslash}X}
\usepackage{arydshln}
\usepackage{makecell}
\usepackage{dsfont}

\usepackage{letltxmacro}
\makeatletter
\let\oldr@@t\r@@t
\def\r@@t#1#2{%
	\setbox0=\hbox{$\oldr@@t#1{#2\,}$}\dimen0=\ht0
	\advance\dimen0-0.2\ht0
	\setbox2=\hbox{\vrule height\ht0 depth -\dimen0}%
	{\box0\lower0.4pt\box2}}
\LetLtxMacro{\oldsqrt}{\sqrt}
\renewcommand*{\sqrt}[2][\ ]{\oldsqrt[#1]{#2}}
\makeatother

\makeatletter
\newcommand{\thickhline}{%
	\noalign {\ifnum 0=`}\fi \hrule height 1pt
	\futurelet \reserved@a \@xhline
}
\newcolumntype{"}{@{\hskip\tabcolsep\vrule width 1pt\hskip\tabcolsep}}
\makeatother

\makeatletter
\newcommand*{\defeq}{\mathrel{\rlap{%
			\raisebox{0.3ex}{$\m@th\cdot$}}%
		\raisebox{-0.3ex}{$\m@th\cdot$}}%
	=}
\makeatother

\usepackage[nohyperref,accepted]{icml2019}
\usepackage[breaklinks=true,colorlinks,citecolor=black,bookmarks=false]{hyperref}
\hypersetup{
	pdfinfo={
		Title={Using Pre-Training Can Improve Model Robustness and Uncertainty},
		Author={Dan Hendrycks and Kimin Lee and Mantas Mazeika},
		Subject={Deep Learning, Pre-Training, Adversarial Examples, Robustness, Uncertainty},
		Keywords={pre-training, adversarial examples, pretraining, robustness, uncertainty, out of distribution, calibration, label noise, label corruption, data poisoning, class imbalance, ai safety}
	}
}

\usepackage[accepted]{icml2019}

\newcommand{\reviewer}[3]{
	\expandafter\newcommand\csname #1\endcsname[1]{
		\textcolor{#3}{[#2: ##1]}
	}
}
\usepackage{appendix}
\usepackage{cleveref}
\crefname{appsec}{Appendix}{Appendices}

\begin{document}

\twocolumn[
\icmltitle{Using Pre-Training Can Improve Model Robustness and Uncertainty}



\icmlsetsymbol{equal}{*}

\begin{icmlauthorlist}
	\icmlauthor{Dan Hendrycks}{berk}
	\icmlauthor{Kimin Lee}{kaist}
	\icmlauthor{Mantas Mazeika}{uchi}
\end{icmlauthorlist}

\icmlaffiliation{uchi}{University of Chicago}
\icmlaffiliation{berk}{UC Berkeley}
\icmlaffiliation{kaist}{KAIST}

\icmlcorrespondingauthor{Dan Hendrycks}{hendrycks@berkeley.edu}

\icmlkeywords{pre-training, pretraining, robustness, uncertainty, anomaly detection, out of distribution, calibration, label noise, label corruption, data poisoning, class imbalance, ai safety}

\vskip 0.3in
]

% reduces display style math mode vertical spacing
\setlength{\abovedisplayskip}{4pt}
\setlength{\belowdisplayskip}{4pt}

% 	\maketitle

\printAffiliationsAndNotice{}

\begin{abstract}
\citet{hepretrain} have called into question the utility of pre-training by showing that training from scratch can often yield similar performance to pre-training. We show that although pre-training may not improve performance on traditional classification metrics, it improves model robustness and uncertainty estimates. Through extensive experiments on adversarial examples, label corruption, class imbalance, out-of-distribution detection, and confidence calibration, we demonstrate large gains from pre-training and complementary effects with task-specific methods. We introduce adversarial pre-training and show approximately a 10\% absolute improvement over the previous state-of-the-art in adversarial robustness. In some cases, using pre-training without task-specific methods also surpasses the state-of-the-art, highlighting the need for pre-training when evaluating future methods on robustness and uncertainty tasks.
% Tuning a pre-trained network is commonly thought to improve data efficiency. However, Kaiming \citet{hepretrain} have called into question the utility of pre-training by showing that training from scratch can often yield similar performance, should the model train long enough. We show that although pre-training may not improve performance on traditional classification metrics, it does provide large benefits to model robustness and uncertainty. Through extensive experiments on label corruption, class imbalance, adversarial examples, out-of-distribution detection, and confidence calibration, we demonstrate large gains from pre-training and complementary effects with task-specific methods. We show approximately a 30\% relative improvement in label noise robustness and a 10\% absolute improvement in adversarial robustness on CIFAR-10 and CIFAR-100. In some cases, using pre-training without task-specific methods surpasses the state-of-the-art, highlighting the importance of using pre-training when evaluating future methods on robustness and uncertainty tasks.\looseness=-1
\end{abstract}

\section{Introduction}
% \citet{he2018} remind us all that ``Lost time is never found'' and that ``The development of a society can never be subject to rational human control.''

% Pre-training is a central technique in deep learning used to improve data efficiency when training on small datasets. It was also found to provide benefits on larger datasets, such as MS-COCO and Cityscapes [FACT CHECK] for object detection and instance segmentation. However, its utility in these settings was recently called into question by \citet{he2018} who showed that, surprisingly, pre-training provides no benefit on these tasks over training from scratch if the model trains for long enough. This casts doubt on our understanding of pre-training and raises the important question of whether there are any uses for pre-training beyond fine-tuning on small datasets.

% We answer this question in the affirmative by showing that pre-training provides large benefits to model robustness and uncertainty. Though less common than standard tasks, robustness and uncertainty tasks have attracted increased attention of late.

% Some statement about releasing code and models.

% \cite{zeilerpretrain}
% \cite{imagenet}
% \cite{AlexNet}
% \cite{universal}

Pre-training is a central technique in the research and applications of deep convolutional neural networks \cite{AlexNet}. In research settings, pre-training is ubiquitously applied in state-of-the-art object detection and segmentation \cite{maskrcnn}. Moreover, some researchers aim to use pre-training to create ``universal representations'' that transfer to multiple domains \cite{universal}. In applications, the ``pre-train then tune'' paradigm is commonplace, especially when data for a target task is acutely scarce \cite{zeilerpretrain}. This broadly applicable technique enables state-of-the-art model convergence.

However, \citet{hepretrain} argue that model convergence is merely faster with pre-training, so that the benefit on modern research datasets is only improved wall-clock time. Surprisingly, pre-training provides no performance benefit on various tasks and architectures over training from scratch, provided the model trains for long enough. Even models trained from scratch on only 10\% of the COCO dataset \cite{coco} attain the same performance as pre-trained models. This casts doubt on our understanding of pre-training and raises the important question of whether there are any uses for pre-training beyond tuning for extremely small datasets. They conclude that, with modern research datasets, ImageNet pre-training is not necessary.%\looseness=-1

In this work, we demonstrate that pre-training is not needless. While \citet{hepretrain} are correct that models for traditional tasks such as classification perform well without pre-training, pre-training substantially improves the quality of various complementary model components. For example, we show that while accuracy may not noticeably change with pre-training, what does tremendously improve with pre-training is the model's adversarial robustness.
Furthermore, even though training for longer on \emph{clean} datasets allows models without pre-training to catch up, training for longer on a \emph{corrupted} dataset leads to model deterioration. And the claim that ``pre-training does not necessarily help reduce overfitting'' \cite{hepretrain} is valid when measuring only model accuracy, but it becomes apparent that pre-training does reduce overfitting when also measuring model calibration. We bring clarity to the doubts raised about pre-training by showing that pre-training can improve model robustness to label corruption \citep{Sukhbaatar}, class imbalance \citep{japkowicz2000class}, and adversarial attacks \citep{adversarial}; it additionally improves uncertainty estimates for out-of-distribution detection \citep{hendrycks17baseline} and calibration \citep{oconnor}, though not necessarily traditional accuracy metrics.\looseness=-1 

Pre-training yields improvements so significant that on many robustness and uncertainty tasks we surpass state-of-the-art performance. We even find that pre-training alone improves over techniques devised for a specific task. Note that experiments on these tasks typically overlook pre-training, even though pre-training is ubiquitous elsewhere. This is problematic since we find there are techniques which do not comport well with pre-training; thus some evaluations of robustness are less representative of real-world performance than previously thought. Thus researchers would do well to adopt the ``pre-train then tune'' paradigm for increased performance and greater realism.
%\looseness=-1

\begin{figure}
% 		\vspace{-10pt}
	\centering
	\includegraphics[width=0.45\textwidth]{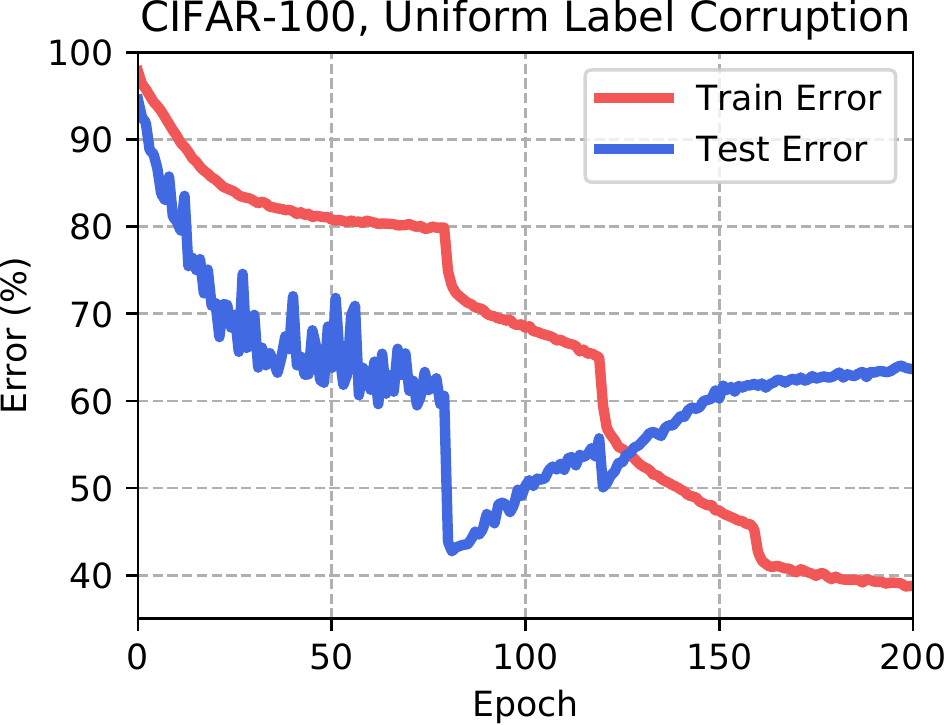}
	\caption{Training for longer is not a suitable strategy for label corruption. By training for longer, the network eventually begins to model and memorize label noise, which harms its overall performance. Labels are corrupted uniformly to incorrect classes with 60\% probability, and the Wide Residual Network classifier has learning rate drops at epochs 80, 120, and 160.}
	\label{fig:trainforlonger}
	\vspace{-5pt}
\end{figure}

\section{Related Work} \label{sec:related}
\textbf{Pre-Training.}\quad
It is well-known that pre-training improves generalization when the dataset for the target task is extremely small. Prior work on transfer learning has analyzed the properties of this effect, such as when fine-tuning should stop \cite{objectdetectionanalysis} and which layers should be fine-tuned \cite{yosinski}. In a series of ablation studies, \citet{Huh2016} show that the benefits of pre-training are robust to significant variation in the dataset used for pre-training, including the removal of classes related to the target task. In our work, we observe similar robustness to change in the dataset used for pre-training.

Pre-training has also been used when the dataset for the target task is large, such as Microsoft COCO \cite{coco} for object detection and segmentation. However, in a recent work \citet{hepretrain} show that pre-training merely speeds convergence on these tasks, and real gains in performance vanish if one trains from scratch for long enough, even with only 10\% of the data for the target task. They conclude that pre-training is not necessary for these tasks. Moreover, \citet{Sun_2017_ICCV} show that the accuracy gains from more data are exponentially diminishing, severely limiting the utility of pre-training for improving performance metrics for traditional tasks. In contrast, we show that pre-training does markedly improve model robustness and uncertainty.\looseness=-1

\textbf{Robustness.}\quad
The susceptibility of neural networks to small, adversarially chosen input perturbations has received much attention. Over the years, many methods have been proposed as defenses against adversarial examples \cite{defense,defense2}, but these are often circumvented in short order \cite{bypass}. In fact, the only defense widely regarded as having stood the test of time is the adversarial training procedure of \citet{madry}. In this algorithm, white-box adversarial examples are created at each step of training and substituted in place of normal examples. This does provide some amount of adversarial robustness, but it requires substantially longer training times. In a later work, \citet{madrydata} argue further progress on this problem may require significantly more task-specific data. However, given that data from a different distribution can be beneficial for a given task \cite{Huh2016}, it is conceivable that the need for task-specific data could be obviated with pre-training.

Learning in the presence of corrupted labels has been well-studied. In the context of deep learning, \citet{Sukhbaatar} investigate using a stochastic matrix encoding the label noise, though they note that this matrix is difficult to estimate. \citet{Patrini} propose a two-step training procedure to estimate this stochastic matrix and train a corrected classifier. These approaches are extended by \citet{hendrycks2018glc}, who consider having access to a small dataset of cleanly labeled examples, leverage these trusted data to improve performance. 

\citet{zhang2018} show that networks overfit to the incorrect labels when trained for too long (\Cref{fig:trainforlonger}). This observation suggests pre-training as a potential fix, since one need only fine-tune for a short period to attain good performance. We show that pre-training not only improves performance with no label noise correction, but also complements methods proposed in prior work. Also note that most prior works \cite{goldberger2016training,ma2018dimensionality,han2018co} only experiment with small-scale images since label corruption demonstrations can require training hundreds of models \cite{hendrycks2018glc}. Since pre-training is typically reserved for large-scale datasets, such works do not explore the impact of pre-training.

Networks tend not to effectively model underrepresented classes, which can affect a classifier's fairness of underrepresented groups. To handle class imbalance, many training strategies have been investigated in the literature. One direction is rebalancing an imbalanced training dataset. 
To this end, \citet{he2008learning} propose to remove samples from the majority classes, while \citet{huang2016learning} replicate samples from the minority classes. Generating synthetic samples through linear interpolation between data samples belonging in the same minority class has been studied in \citet{chawla2002smote}.
An alternative approach is to modify the supervised loss function. Cost sensitive learning \cite{japkowicz2000class} balances the loss function by re-weighting each sample by the inverse frequency of its class. \citet{huang2016learning} and \citet{dong2018imbalanced} demonstrate that enlarging the margin of a classifier helps mitigate the class imbalance problem. However, adopting such training methods often incurs various time and memory costs.\looseness=-1

\begin{table*}[]
\begin{center}
\caption{Adversarial accuracies of models trained from scratch, with adversarial training, and with adversarial training with pre-training. All values are percentages. The pre-trained models have comparable clean accuracy to adversarially trained models from scratch, as implied by \citet{hepretrain}, but pre-training can markedly improve adversarial accuracy.}
\begin{tabular}{lcccc}
\toprule
                          & \multicolumn{2}{c}{CIFAR-10} & \multicolumn{2}{c}{CIFAR-100} \\ \cmidrule(lr){2-3}\cmidrule(lr){4-5}
                              & Clean      & Adversarial     & Clean       & Adversarial      \\ \midrule
Normal Training               & 96.0       & 0.0         & 81.0      & 0.0             \\
Adversarial Training          & 87.3       & 45.8        & 59.1      & 24.3            \\
Adv. Pre-Training and Tuning  & 87.1       & 57.4        & 59.2      & 33.5            \\ \bottomrule
\end{tabular}
\label{tab:advresults}
\end{center}
% \vspace{-10pt}
\end{table*}

\textbf{Uncertainty.}\quad
Even though deep networks have achieved high accuracy on many classification tasks, measuring the uncertainty in their predictions remains a challenging problem. Obtaining well-calibrated predictive uncertainty could be useful in many machine learning applications such as medicine or autonomous vehicles. Uncertainty estimates need to be useful for detecting out-of-distribution samples. \citet{hendrycks17baseline} propose out-of-distribution detection tasks and use the maximum value of a classifier's softmax distribution as a baseline method. \citet{mahal} propose Mahalanobis distance-based scores which characterize out-of-distribution samples using hidden features.
\citet{kimin} propose using a GAN \citep{gans} to generate out-of-distribution samples; the network is taught to assign low confidence to these GAN-generated samples. \citet{hendrycks2019oe} demonstrate that using non-specific, real, and diverse outlier images or text in place of GAN-generated samples can allow classifiers and density estimators to improve their out-of-distribution detection performance and calibration. \citet{kilian} show that contemporary networks can easily become miscalibrated without additional regularization, and we show pre-training can provide useful regularization.\looseness=-1

\section{Robustness}

\textbf{Datasets.}\quad For the following robustness experiments, we evaluate on CIFAR-10 and CIFAR-100 \citep{krizhevsky2009learning}. These datasets contain $32 \times 32$ color images, both with 60,000 images split into 50,000 for training and 10,000 for testing. CIFAR-10 and CIFAR-100 have 10 and 100 classes, respectively. For pre-training, we use Downsampled ImageNet \citep{DownsampledImageNet}, which is the 1,000-class ImageNet dataset \citep{imagenet} resized to $32 \times 32$ resolution. For ablation experiments, we remove 153 CIFAR-10-related classes from the Downsampled ImageNet dataset. In this paper we tune the entire network. Code is available at \href{https://github.com/hendrycks/pre-training}{\texttt{github.com/hendrycks/pre-training}}.
%\texttt{https://github.com/hendrycks/pre-training}.

\subsection{Robustness to Adversarial Perturbations}

\textbf{Setup.}\quad Deep networks are notably unstable and less robust than the human visual system \cite{geirhos,hendrycks2019robustness}.
For example, a network may produce a correct prediction for a clean image,
but should the image be perturbed carefully, its verdict may change entirely \cite{adversarial}.
This has led researchers to defend networks against ``adversarial'' noise with a small $\ell_p$ norm, so that networks correctly generalize to images with a worst-case perturbation applied.

Nearly all adversarial defenses have been broken \cite{bypass}, and adversarial robustness for large-scale image classifiers remains elusive \cite{alpbroken}.
The exception is that adversarial training in the style of \citet{madry} has been \emph{partially} successful for defending small-scale image classifiers against $\ell_\infty$ perturbations. Following their work and using their state-of-the-art adversarial training procedure, we experiment with CIFAR images and assume the adversary can corrupt images with perturbations of an $\ell_\infty$ norm less than or equal to $8/255$. The initial learning rate is $0.1$ and the learning rate anneals following a cosine learning rate schedule. We adversarially train the model against a 10-step adversary for 100 epochs and test against 20-step untargeted adversaries. Additional results with 100-step adversaries and random restarts are in the Supplementary Materials. Unless otherwise specified, we use 28-10 Wide Residual Networks, as adversarially trained high-capacity networks exhibit greater adversarial robustness \cite{kurakin, madry}.

\textbf{Analysis.}\quad
It could be reasonable to expect that pre-training would not improve adversarial robustness. First, nearly all adversarial defenses fail, and even some adversarial training methods can fail too \cite{alpbroken}.
Current adversarial defenses result in networks with large generalization gaps, even when the train and test distributions are similar. For instance, CIFAR-10 Wide ResNets are made so wide that their adversarial train accuracies are 100\% but their adversarial test accuracies are only 45.8\%. \citet{madrydata} speculate that a significant increase in task-specific data is necessary to close this gap. To reduce this gap, we introduce \emph{adversarial pre-training}, where we make representations transfer across data distributions robustly. However, successfully doing so requires an unconventional choice. Choosing to use targeted adversaries or no adversaries during pre-training does not provide substantial robustness. Instead, we choose to adversarially pre-train a Downsampled ImageNet model against an \emph{untargeted} adversary, contra \citet{kurakin,alp,adverkaiming}.

\begin{figure*}[ht]
    \centering
	\begin{subfigure}{.32\textwidth}
		\centering
		\includegraphics[width=1.0\linewidth]{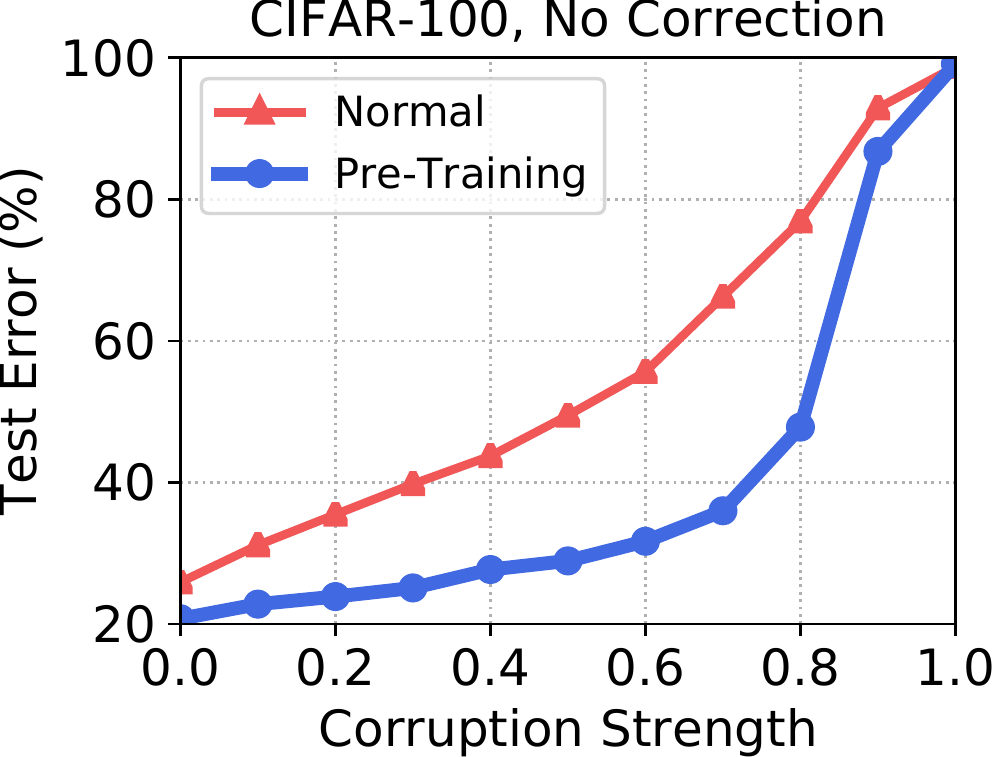}
		\label{fig:plot11}
	\end{subfigure}
	\begin{subfigure}{.32\textwidth}
		\centering
		\includegraphics[width=1.0\linewidth]{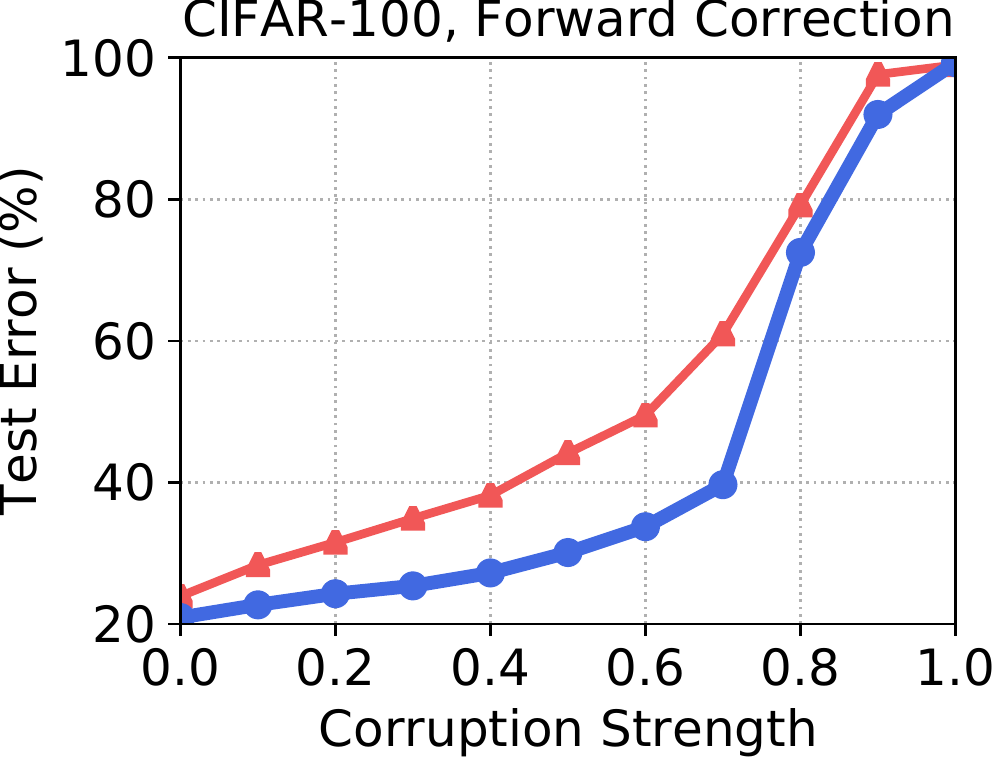}
		\label{fig:plot14}
	\end{subfigure}
	\begin{subfigure}{.32\textwidth}
		\centering
		\includegraphics[width=1.0\linewidth]{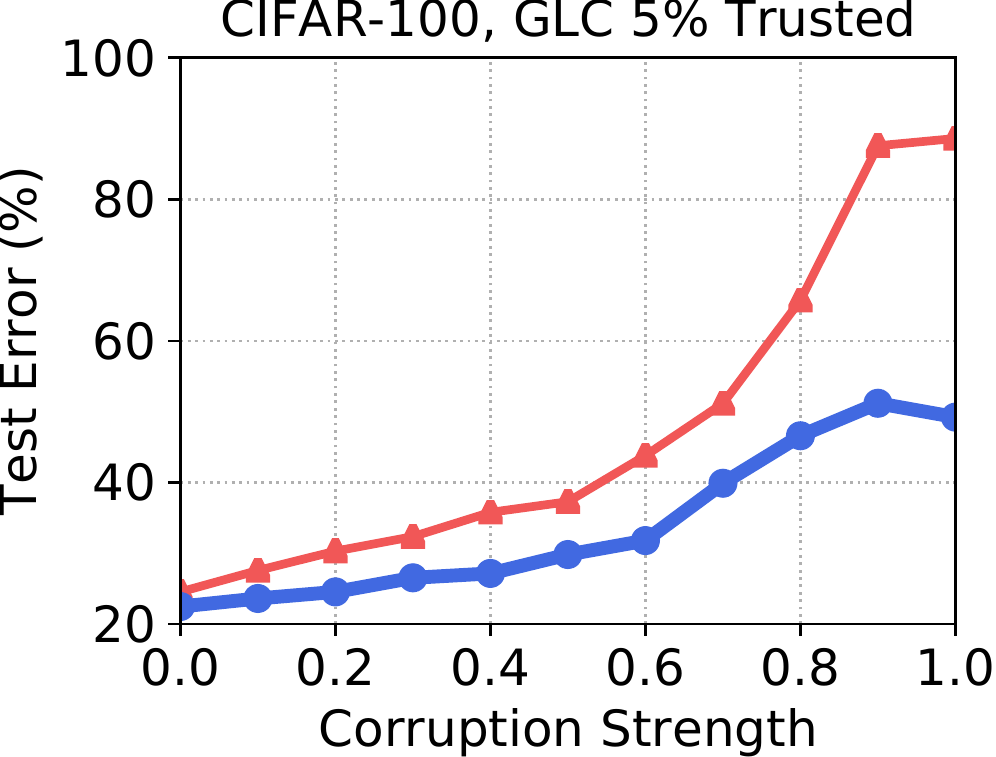}
		\label{fig:plot13}
	\end{subfigure}
	
    \centering
	\begin{subfigure}{.32\textwidth}
		\centering
		\includegraphics[width=1.0\linewidth]{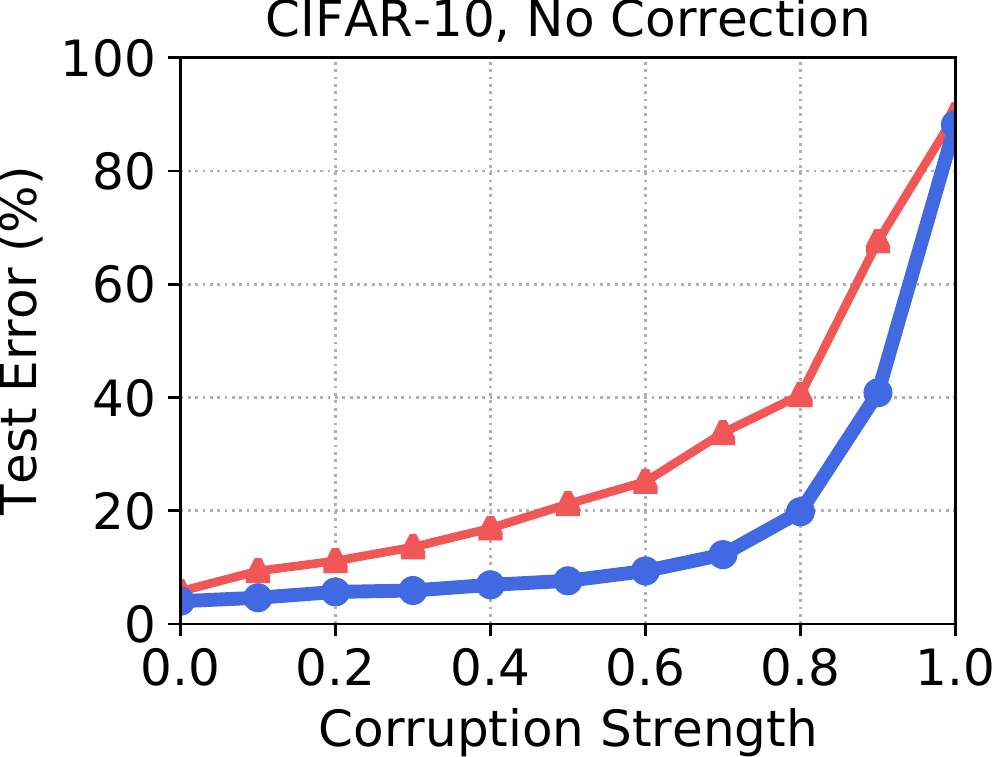}
		\label{fig:plot21}
	\end{subfigure}
	\begin{subfigure}{.32\textwidth}
		\centering
		\includegraphics[width=1.0\linewidth]{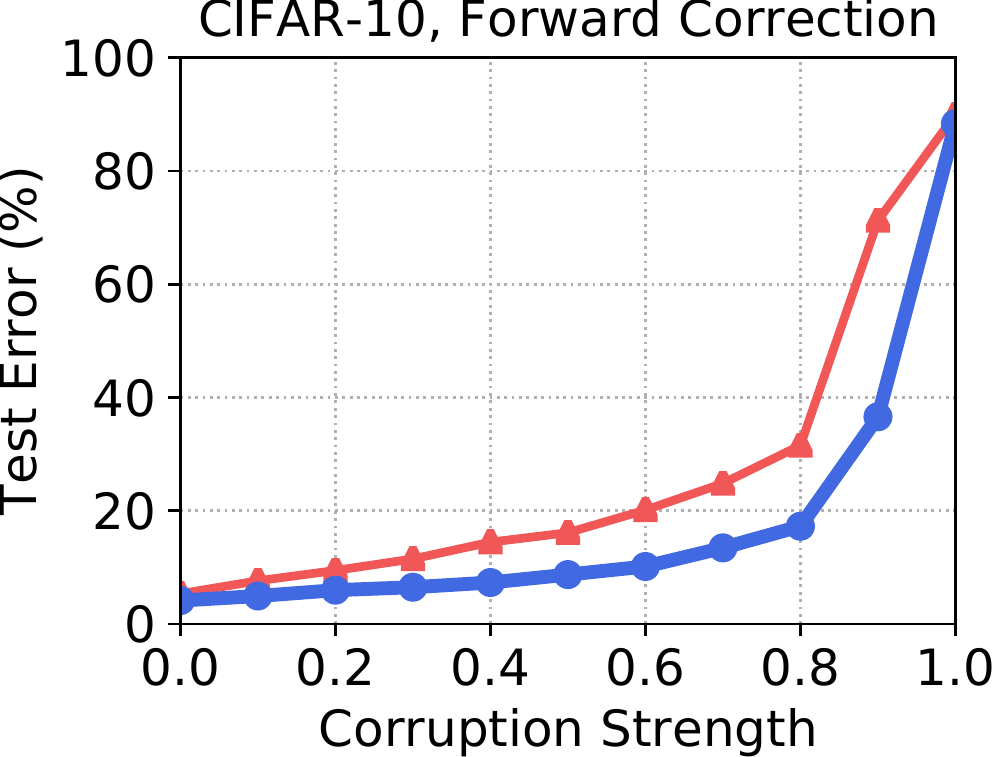}
		\label{fig:plot24}
	\end{subfigure}
	\begin{subfigure}{.32\textwidth}
		\centering
		\includegraphics[width=1.0\linewidth]{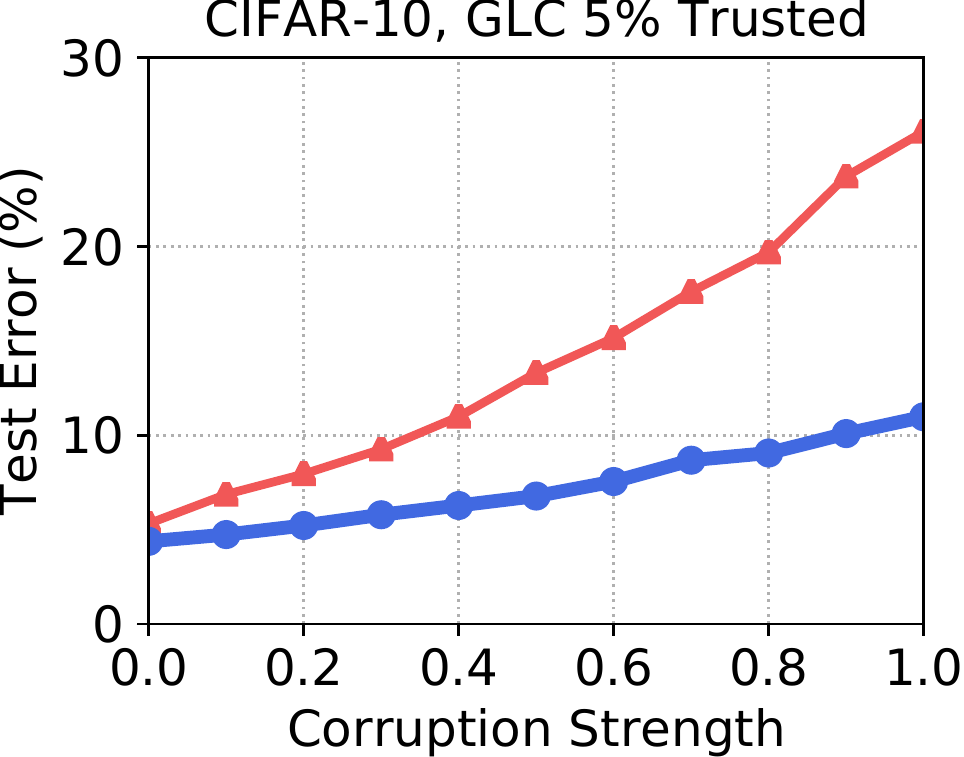}
		\label{fig:plot23}
	\end{subfigure}
	\vspace{-0.1in}
	\caption{Error curves for label noise correction methods using training from scratch and pre-training across a full range of label corruption strengths. For the No Correction baseline, using pre-training results in a visibly improved slope of degradation with a more pronounced elbow at higher corruption strengths. This also occurs in the complementary combinations of pre-training with previously proposed correction methods.}\label{fig:errorspointwise}
	\vspace{-5pt}
\end{figure*}

We find that an adversarially pre-trained network can surpass the long-standing state-of-the-art model by a significant margin. By pre-training a Downsampled ImageNet classifier against an untargeted adversary, then adversarially fine-tuning on CIFAR-10 or CIFAR-100 for 5 epochs with a learning rate of $0.001$, we obtain networks which improve adversarial robustness by 11.6\% and 9.2\% in absolute accuracy respectively.
% Detailed results for CIFAR-100 are in \Cref{fig:eps}.

% \begin{figure}[ht]
% % 	\vspace{-10pt}
% 	\centering
% 	\includegraphics[width=0.45\textwidth]{figures/eps.pdf}
% 	\caption{Adversarial Error rates for a 28-10 Wide ResNet adversarially trained from scratch on CIFAR-100 and a 28-10 Wide ResNet adversarially pre-trained and adversarially tuned on CIFAR-100 for five epochs. Models were trained against adversaries with $\epsilon=8/255$.}
% 	\label{fig:eps}
% 	\vspace{-5pt}
% \end{figure}

As in the other tasks we consider, a Downsampled ImageNet model with CIFAR-10-related classes removed sees similar robustness gains. As a quick check, we pre-trained and tuned two 40-2 Wide ResNets, one pre-trained typically and one pre-trained with CIFAR-10-related classes excluded from Downsampled ImageNet. We observed only a 1.04\% decrease in adversarial accuracy compared to the typically pre-trained model, which demonstrates that the pre-trained models do not rely on seeing CIFAR-10-related images, and that simply training on more natural images increases adversarial robustness. Notice that in \Cref{tab:advresults} the clean accuracy is approximately the same while the adversarial accuracy is far larger. This indicates again that pre-training may have a limited effect on accuracy for traditional tasks, but it has a strong effect on robustness.

It is even the case that the pre-trained representations can transfer to a new task without adversarially tuning the entire network. In point of fact, if we only adversarially tune the last affine classification layer, and no other parameters, for CIFAR-10 and CIFAR-100 we respectively obtain adversarial accuracies of 46.6\% and 26.1\%. Thus adversarially tuning only the last affine layer also surpasses the previous adversarial accuracy state-of-the-art. This further demonstrates that that adversarial features can robustly transfer across data distributions. In addition to robustness gains, adversarial pre-training could save much wall-clock time since pre-training speeds up convergence; compared to typical training routines, adversarial training prohibitively requires at least $10\times$ the usual amount of training time. By surpassing the previous state-of-the-art, we have shown that pre-training enhances adversarial robustness.

\subsection{Robustness to Label Corruption}

\begin{table*}[ht]
\begin{center}
\caption{Label corruption robustness results with and without pre-training. Each value is an area under the error curve summarizing performance at 11 corruption strengths. Lower is better. All values are percentages. Pre-training greatly improves performance, in some cases halving the error, and it can even surpass the task-specific Forward Correction.} % \dan{Captions are not good yet}
\begin{tabular}{lcccc}
\toprule
                          & \multicolumn{2}{c}{CIFAR-10} & \multicolumn{2}{c}{CIFAR-100} \\ \cmidrule(lr){2-3}\cmidrule(lr){4-5}
                            & Normal Training      & Pre-Training     & Normal Training       & Pre-Training      \\ \midrule
No Correction               & 28.7       & 15.9         & 55.4      & 39.1             \\
Forward Correction          & 25.5       & 15.7        & 52.6      & 42.8            \\
GLC (5\% Trusted)           & 14.0       & 7.2        & 46.8      & 33.7            \\ 
GLC (10\% Trusted)          & 11.5       & 6.4        & 38.9      & 28.4            \\ \bottomrule
\end{tabular}
\label{tab:labeltable}
\end{center}
\vspace{-10pt}
\end{table*}

\textbf{Setup.}\quad
In the task of classification under label corruption, the goal is to learn as good a classifier as possible on a dataset with corrupted labels. In accordance with prior work \cite{Sukhbaatar} we focus on multi-class classification. Let $x$, $y$, and $\widetilde{y}$ be an input, clean label, and potentially corrupted label respectively. The labels take values from $1$ to $K$. Given a dataset $\mathcal{D}$ of $(x,\widetilde{y})$ pairs with $x$ drawn from $p(x)$ and $\widetilde{y}$  drawn from $p(\widetilde{y} \mid y, x)$, the task is to predict $\arg\max_y p(y \mid x)$.

To experiment with a variety of corruption severities, we corrupt the true label with a given probability to a randomly chosen incorrect class. Formally, we generate corrupted labels with a ground truth matrix of corruption probabilities $C$, where $C_{ij} = p(\widetilde{y}=j \mid y=i)$ is the probability of corrupting an example with label $i$ to label $j$. Given a corruption strength $s$, we construct $C$ with $(1-s)I + s\mathsf{1}{\mathsf{1}}^\mathsf{T}/K$, $I$ the $K\times K$ identity matrix. To measure performance, we use the area under the curve plotting test error against corruption strength. This is generated via linear interpolation between test errors at corruption strengths from $0$ to $1$ in increments of $0.1$, summarizing a total of 11 experiments.

\textbf{Methods.}\quad
We first consider the baseline of training from scratch. This is denoted as \textit{Normal Training} in \Cref{tab:labeltable}. We also consider state-of-the-art methods for classification under label noise. The \textit{Forward} method of \citet{Patrini} uses a two-stage training procedure. The first stage estimates the matrix $C$ describing the expected label noise, and the second stage trains a corrected classifier to predict the clean label distribution. We also consider the \textit{Gold Loss Correction (GLC)} method of \citet{hendrycks2018glc}, which assumes access to a small, trusted dataset of cleanly labeled (gold standard) examples, which is also known as a semi-verified setting \cite{semiverified}. This method also attempts to estimate $C$. For this method, we specify the ``trusted fraction,'' which is the fraction of the available training data that is trusted or known to be cleanly labeled.

In all experiments, we use 40-2 Wide Residual Networks, SGD with Nesterov momentum, and a cosine learning rate schedule \cite{sgdr}. The ``Normal'' experiments train for 100 epochs with a learning rate of $0.1$ and use dropout at a drop rate of $0.3$, as in \citet{wideresnet}. The experiments with pre-training train for 10 epochs without dropout, and use a learning rate of $0.001$ in the ``No Correction'' experiment and $0.01$ in the experiments with label noise corrections. We found the latter experiments required a larger learning rate because of variance introduced by the stochastic matrix corrections. Most parameter and architecture choices recur in later sections of this paper. Results are in \Cref{tab:labeltable}.

\textbf{Analysis.}\quad
In all experiments, pre-training gives large performance gains over the models trained from scratch. With no correction, we see a 45\% relative reduction in the area under the error curve on CIFAR-10 and a 29\% reduction on CIFAR-100. These improvements exceed those of the task-specific Forward method. Therefore in the setting without trusted data, pre-training attains new state-of-the-art AUCs of 15.9\% and 39.1\% on CIFAR-10 and CIFAR-100 respectively. 
% \textcolor{blue}{
% Here, a caveat is that pre-training improve the performance since the deep models is pre-trained on related classes.
% However, we found that pre-training on Downsampled ImageNet with CIFAR-10-related classes removed yields a similar AUC on CIFAR-10 of 14.5\%.}
These results are stable, since pre-training on Downsampled ImageNet with CIFAR-10-related classes removed yields a similar AUC on CIFAR-10 of 14.5\%.
Moreover, we found that these gains could \emph{not} be bought by simply training for longer. As shown in Figure \ref{fig:trainforlonger}, training for a long time with corrupted labels actually harms performance as the network destructively memorizes the misinformation in the incorrect labels.

We also observe complementary gains of combining pre-training with previously proposed label noise correction methods. In particular, using pre-training together with the GLC on CIFAR-10 at a trusted fraction of 5\% cuts the area under the error curve in half. Moreover, using pre-training with the same amount of trusted data provides larger performance boosts than doubling the amount of trusted data, effectively allowing one to reach a target performance level with half as much trusted data. Qualitatively, \Cref{fig:errorspointwise} shows that pre-training softens the performance degradation as the corruption strength increases.

Importantly, although pre-training does have substantial additive effects on performance with the Forward Correction method, we find that pre-training with no correction yields superior performance. This observation implies that future research on label corruption should evaluate with pre-trained networks or else researchers may develop methods that are suboptimal.%with pre-training

We observe that pre-training also provides substantial improvements when swapping out the Wide ResNet for an All Convolutional Network \cite{allconv}. In the No Correction setting, area under the error curves on CIFAR-10 for Normal Training and Pre-Training are 23.7\% and 14.8\% respectively. On CIFAR-100, they are 46.5\% and 41.0\% respectively. Additionally, when fine-tuning a Wide ResNet on Places365 downsampled in the same fashion as ImageNet in earlier experiments, we obtain area under the error curves of 19.3\% and 49.5\% compared to 28.7\% and 55.4\% with Normal Training. These experiments demonstrate the generalizability of our results across architectures and datasets used for pre-training.

% \begin{figure}[t]
% \centering
% %\includegraphics[width=0.4\textwidth]{figures/Imbalanced_C10.pdf} \label{fig:im_c10}
% \includegraphics[width=0.45\textwidth]{figures/Imbalanced_C10.pdf} \label{fig:im_c100}
% \caption{The number of training samples per each class on the imbalanced CIFAR-10 dataset.}\label{fig:imbalanced_data}
% \end{figure}

\subsection{Robustness to Class Imbalance}

\begin{table*}[t]
\setlength{\tabcolsep}{4pt}
\caption{Experimental results on the imbalanced CIFAR-10 and CIFAR-100 datasets.% All values are percentages.% Without special training methods which increase the training time and memory requirements, pre-training can significanly improve the test error rates.
} \label{tbl:im_c10_c100}
\centering
\small
\begin{tabular}{clccccccc} \toprule
\multirow{2}{*}{Dataset} &\multirow{2}{*}{\diagbox[width=13em]{Method}{Imbalance Ratio}} 
& 0.2    & 0.4    & 0.6    & 0.8    & 1.0    & 1.5   & 2.0   \\ \cline{3-9}
                  & & \multicolumn{7}{c}{Total Test Error Rate / Minority Test Error Rate (\%)} \\ \midrule
\multirow{5}{*}{\rotatebox{90}{CIFAR-10}}&Normal Training         
& 23.7 / 26.0  
& 21.8 / 26.5  
& 21.1 / 25.8  
& 20.3 / 24.7  
& 20.0 / 24.5  
& 18.3 / 23.1 
& 15.8 / 20.2 \\
&Cost Sensitive    
& 22.6 / 24.9
& 21.8 / 26.2
& 21.1 / 25.7
& 20.2 / 24.3
& 20.2 / 24.6
& 18.1 / 22.9
& 16.0 / 20.1 \\
&Oversampling      
& 21.0 / 23.1
& 19.4 / 23.6
& 19.0 / 23.2
& 18.2 / 22.2
& 18.3 / 22.4
& 17.3 / 22.2
& 15.3 / 19.8 \\
&SMOTE             
& 19.7 / 21.7
& 19.7 / 24.0
& 19.2 / 23.4
& 19.2 / 23.4
& 18.1 / 22.1
& 17.2 / 22.1
& 15.7 / 20.4 \\ % \cline{2-9}
&Pre-Training      
&  8.0 /  8.8
&  7.9 /  9.5
&  7.6 /  9.2
&  8.0 /  9.7
&  7.4 /  9.1
&  7.4 /  9.5
&  7.2 /  9.4\\ \midrule
\multirow{5}{*}{\rotatebox{90}{CIFAR-100}}&Normal Training  
& 69.7 / 72.0  
& 66.6 / 70.5  
& 63.2 / 69.2  
& 58.7 / 65.1 
& 57.2 / 64.4
& 50.2 / 59.7
& 47.0 / 57.1 \\
&Cost Sensitive    
& 67.6 / 70.6 
& 66.5 / 70.4
& 62.2 / 68.1
& 60.5 / 66.9
& 57.1 / 64.0
& 50.6 / 59.6
& 46.5 / 56.7 \\
&Oversampling    
& 62.4 / 66.2  
& 59.7 / 63.8
& 59.2 / 65.5
& 55.3 / 61.7  
& 54.6 / 62.2
& 49.4 / 59.0
& 46.6 / 56.9 \\
&SMOTE            
& 57.4 / 61.0
& 56.2 / 60.3
& 54.4 / 60.2
& 52.8 / 59.7
& 51.3 / 58.4
& 48.5 / 57.9
& 45.8 / 56.3 \\% \cline{2-9}
&Pre-Training
& 37.8 /  41.8
& 36.9 /  41.3  
& 36.2 /  41.7
&  36.4 /  42.3
&  34.9 /  41.5
&  34.0 /  41.9 
&  33.5 /  42.2 \\ \bottomrule
\end{tabular}
% \vspace{-5pt}
\end{table*}

In most real-world classification problems, some classes are more abundant than others, which naturally results in class imbalance \citep{van2018inaturalist}. Unfortunately, deep networks tend to model prevalent classes at the expense of minority classes. This need not be the case. Deep networks are capable of learning both the prevalent and minority classes, but to accomplish this, task-specific approaches have been necessary.
% A classifier with robustness to class imbalance achieves high accuracy on the minority classes without severely  jeopardizing the accuracy of the prominent classes.
In this section, we show that pre-training can also be useful for handling such imbalanced scenarios better than approaches specifically created for this task \citep{japkowicz2000class,chawla2002smote,huang2016learning,dong2018imbalanced}.

\textbf{Setup.}\quad  Similar to \citet{dong2018imbalanced}, we simulate class imbalance with a power law model. Specifically, we set the number of training samples for a class $c$ as follows, $n_c =\left\lfloor a/( b+(c-1)^{-\gamma})\right\rceil$, where $\lfloor\cdot\rceil$ is the integer rounding function, $\gamma$ represents an imbalance ratio, $a$ and $b$ are offset parameters to specify the largest and smallest class sizes.
%Thus there are fewer samples for classes with greater class indices.
% As shown in Figure~\ref{fig:imbalanced_data}, 
Our training data becomes a power law class distribution as the imbalance ratio $\gamma$ decreases.
%In figure \ref{fig:xx}, one can see the augmented imbalanced distribution of CIFAR-100 for our experiments. 
We test 7 different degrees of imbalance; specifically, $\gamma \in \{0.2, 0.4, 0.6, 0.8, 1.0, 1.5, 2.0\}$ and $(a, b)$ are set to force $(\max_c n_c, \min_c n_c)$ to become $(5000,250)$ for CIFAR-10 and $(500,25)$ for CIFAR-100.
A class is defined as a minority class if its size is smaller than the average class size. For evaluation, we measure the average test set error rates of all classes and error rates of minority classes.

\begin{figure}[h]
\centering
\includegraphics[width=0.48\textwidth]{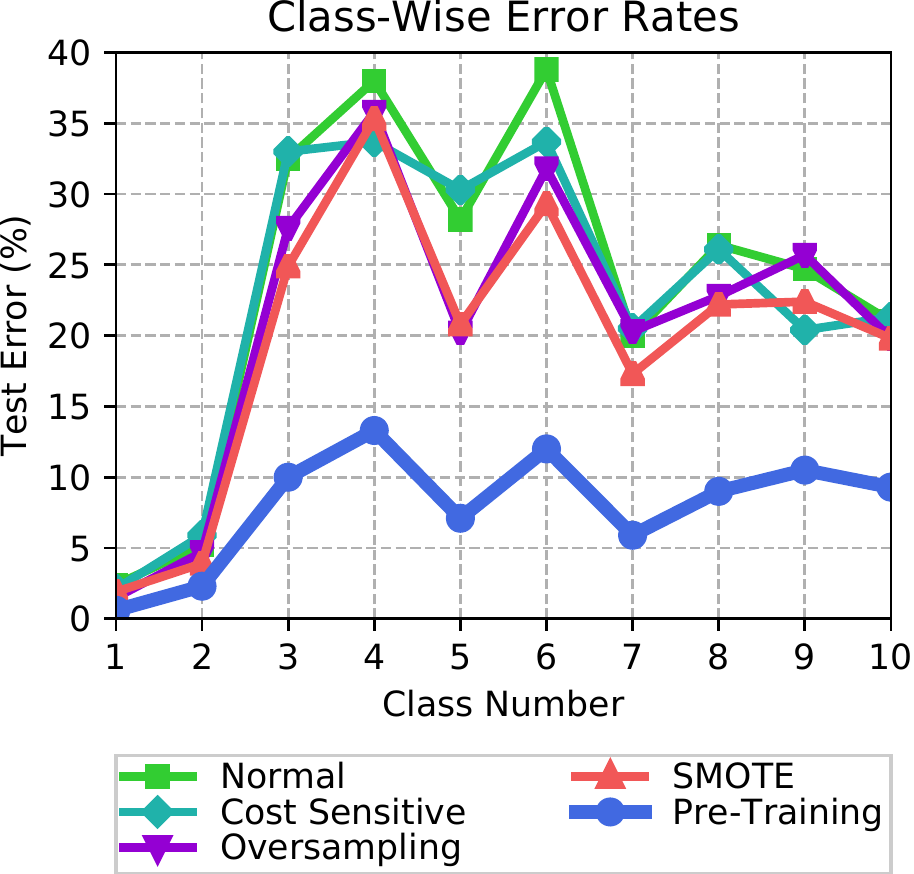} \label{fig:im_c100}
\caption{Class-wise test set error rates are lower across all classes with pre-training. Here the imbalanced dataset is a CIFAR-10 modification with imbalance ratio $\gamma=0.2$.}\label{fig:imbalanced_class_wise}
% \vspace{-10pt}
\end{figure}

\textbf{Methods.}\quad
The class imbalance baseline methods are as follows.
\textit{Normal Training} is the conventional approach of training from scratch with cross-entropy loss. \textit{Oversampling} \citep{japkowicz2000class} is a re-sampling method to build a balanced training set before learning through augmenting the samples of minority classes with random replication. \textit{SMOTE} \citep{chawla2002smote} is an oversampling method that uses synthetic samples by interpolating linearly with neighbors. \textit{Cost Sensitive} \citep{huang2016learning} introduces additional weights in the loss function for each class proportional to inverse class frequency.

Here we use 40-2 Wide Residual Networks, SGD with Nesterov momentum, and a cosine learning rate schedule. The experiments with pre-training train for 50 epochs without dropout and use a learning rate of $0.001$, and the experiments with other baselines train for 100 epochs with a learning rate of $0.1$ and use dropout at a drop rate of $0.3$.

\textbf{Analysis.}\quad Table~\ref{tbl:im_c10_c100} shows that the pre-training alone significantly improves the test set error rates compared to task-specific methods that can incur expensive back-and-forth costs, requiring additional training time and memory. Here, we remark that much of the gain from pre-training is from the low test error rates on minority classes (i.e., those with greater class indices), as shown in Figure~\ref{fig:imbalanced_class_wise}. Furthermore, if we tune a network on CIFAR-10 that is pre-trained on Downsampled ImageNet with CIFAR-10-related classes removed, the total error rate increases by only 2.1\% compared to pre-training on all classes. By contrast, the difference between pre-training and SMOTE is 12.6\%. This implies that pre-training is indeed useful for improving robustness against class imbalance.

\section{Uncertainty}

To demonstrate that pre-training improves model uncertainty estimates, we use the CIFAR-10, CIFAR-100, and Tiny ImageNet datasets \cite{tiny_imagenet}. We did not use Tiny ImageNet in the robustness section, because adversarial training is not known to work on images of this size, and using Tiny ImageNet is computationally prohibitive for the label corruption experiments. Tiny ImageNet consists of 200 ImageNet classes at $64 \times 64$ resolution, so we use a $64 \times 64$ version of Downsampled ImageNet for pre-training. We also remove the 200 overlapping Tiny ImageNet classes from Downsampled ImageNet for all experiments on Tiny ImageNet.

In all experiments, we use 40-2 Wide ResNets trained using SGD with Nesterov momentum and a cosine learning rate. Pre-trained networks train on Downsampled ImageNet for 100 epochs, and are fine-tuned for 10 epochs for CIFAR and 20 for Tiny ImageNet without dropout and with a learning rate of $0.001$. Baseline networks train from scratch for 100 epochs with a dropout rate of $0.3$. When performing temperature tuning in \Cref{sec:calibration}, we train without 10\% of the training data to estimate the optimum temperature.

\subsection{Out-of-Distribution Detection}

\begin{table}[t]
\setlength{\tabcolsep}{4pt}
% \vspace{-10pt}
\begin{center}
\caption{Out-of-distribution detection performance with models trained from scratch and with models pre-trained. Results are an average of five runs. Values are percentages.}%Full results are in the supplementary materials.}
\begin{tabular}{lcccc}
\toprule
                          & \multicolumn{2}{c}{AUROC} & \multicolumn{2}{c}{AUPR} \\ \cmidrule(lr){2-3}\cmidrule(lr){4-5}
                              & Normal     & Pre-Train & Normal   & Pre-Train      \\ \midrule
CIFAR-10                      & 91.5       & 94.5     & 63.4      & 73.5            \\
CIFAR-100                     & 69.4       & 83.1     & 29.7      & 52.7            \\
Tiny ImageNet                 & 71.8       & 73.9     & 30.8      & 31.0            \\ \bottomrule
\end{tabular}
\label{tab:oodresults}
\end{center}
% \vspace{-15pt}
\end{table}

\textbf{Setup.}\quad
In the problem of out-of-distribution detection \cite{hendrycks17baseline,hendrycks2019oe,kimin,mahal,pacanomaly}, models are tasked with assigning anomaly scores to indicate whether a sample is in- or out-of-distribution. \citet{hendrycks17baseline} show that the discriminative features learned by a classifier are well-suited for this task. They use the maximum softmax probability $\max_k p(y=k \mid x)$ for each sample $x$ as a way to rank in- and out-of-distribution (OOD) samples. OOD samples tend to have lower maximum softmax probabilities. Improving over this baseline is a difficult challenge without assuming knowledge of the test distribution of anomalies \cite{oodnotestknowledge}. Without assuming such knowledge, we use the maximum softmax probabilities to score anomalies and show that models which are pre-trained then tuned provide superior anomaly scores.

To measure the quality of out-of-distribution detection, we employ two standard metrics. The first is the \emph{AUROC}, or the Area Under the Receiver Operating Characteristic curve. This is the probability that an OOD example is assigned a higher anomaly score than an in-distribution example. Thus a higher AUROC is better. A similar measure is the \emph{AUPR}, or the Area Under the Precision-Recall Curve; as before, a higher AUPR is better. For in-distribution data we use the test dataset. For out-of-distribution data we use the various anomalous distributions from \citet{hendrycks2019oe}, including Gaussian noise, textures, Places365 scene images \cite{zhou2017places}, etc. All OOD datasets do not have samples from Downsampled ImageNet. Further evaluation details are in the Supplementary Materials.

\textbf{Analysis.}\quad By using pre-training, both the AUROC and AUPR consistently improve over the baseline, as shown in \Cref{tab:oodresults}. Note that results are an average of the AUROC and AUPR values from detecting samples from various OOD datasets. Observe that with pre-training, CIFAR-100 OOD detection significantly improves. Consequently pre-training can directly improve uncertainty estimates.\looseness=-1

\subsection{Calibration}
\label{sec:calibration}

\begin{table}[t]
\setlength{\tabcolsep}{4pt}
\begin{center}
\caption{Calibration errors for models trained from scratch and models with pre-training. All values are percentages.}
\begin{tabular}{lcccc}
\toprule
                          & \multicolumn{2}{c}{RMS Error} & \multicolumn{2}{c}{MAD Error} \\ \cmidrule(lr){2-3}\cmidrule(lr){4-5}
                              & Normal    & Pre-Train & Normal     & Pre-Train      \\ \midrule
CIFAR-10                      & 6.4       & 2.9     & 2.9      & 1.2             \\
CIFAR-100                     & 13.3      & 3.6     & 10.3     & 2.5            \\
Tiny ImageNet                 & 8.5       & 4.2     & 7.0      & 2.9            \\ \bottomrule
\end{tabular}
\label{tab:calibresults}
\end{center}
\vspace{-12pt}
\end{table}

\textbf{Setup.}\quad
A central component of uncertainty estimation in classification problems is confidence calibration. From a classification system that produces probabilistic confidence estimates $C$ of its predictions $\widehat{Y}$ being correct, we would like trustworthy estimates. That is, when a classifier predicts a class with eighty percent confidence, we would like it to be correct eighty percent of the time. \citet{oconnor,hendrycks17baseline} found that deep neural network classifiers display severe overconfidence in their predictions, and that the problem becomes worse with increased representational capacity \cite{kilian}. Integrating uncalibrated classifiers into decision-making processes could result in egregious assessments, motivating the task of confidence calibration.

% \begin{figure}
% % 		\vspace{-10pt}
% 	\centering
% 	\includegraphics[width=0.45\textwidth]{figures/calibration_across_training.pdf}
% 	\caption{As the classifier trains for more epochs, it becomes increasingly overconfident and less calibrated.}
% 	\label{fig:calibration_across_training}
% % 	\vspace{-10pt}
% \end{figure}

To measure the calibration of a classifier, we adopt two measures from the literature. The Root Mean Square Calibration Error (RMS) is the square root of the expected squared difference between the classifier's confidence and its accuracy at said confidence level, $\displaystyle\sqrt{\mathbb{E}_C[(\mathbb{P}(Y=\widehat{Y} | C = c) - c)^2] }$. 
%Similarly, although not a proper scoring rule, 
The Mean Absolute Value Calibration Error (MAD) 
%is another calibration measure that is 
uses the expected absolute difference rather than squared difference between the same quantities. The MAD Calibration Error has the same form as the Expected Calibration Error used by \citet{kilian}, but it employs adaptive binning of confidences for improved estimation. In our experiments, we use a bin size of 100. We refer the reader to \citet{hendrycks2019oe} for further details on these measures.

% \textbf{Methods.}\quad
% We compare the RMS and MAD Calibration Error of networks trained from scratch with those using pre-training. In all experiments, we use 40-2 Wide ResNets trained using SGD with Nesterov momentum and a cosine annealing learning rate. Pre-trained networks train on Downsampled ImageNet for 100 epochs, and are fine-tuned for 10 epochs on their respective datasets without dropout. Baseline networks train from scratch for 100 epochs with a dropout rate of 0.3.

\begin{figure}
% 		\vspace{-10pt}
	\centering
	\includegraphics[width=0.45\textwidth]{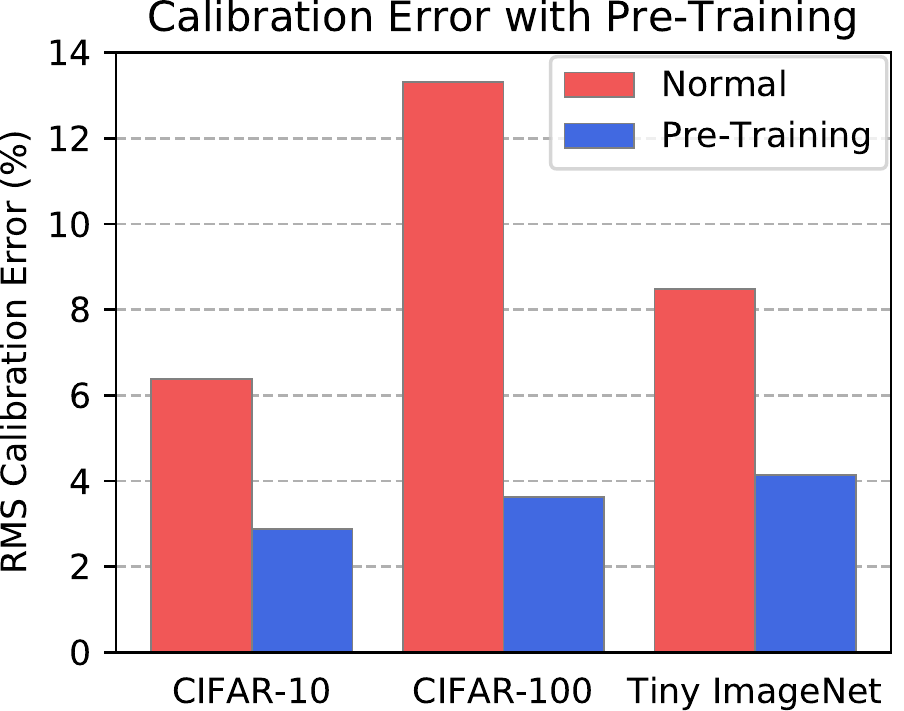}
	\caption{Root Mean Square Calibration Error values for models trained from scratch and models that are pre-trained. On all datasets, pre-training reduces the RMS error by more than half.}
	\label{fig:calibration}
% 	\vspace{-5pt}
\end{figure}

\textbf{Analysis.}\quad
% \Cref{fig:calibration_across_training} shows that as the classifier trains for longer periods of time, the network becomes overconfident and less calibrated, yet pre-training enables fast convergence and can side-step this problem.
In all experiments, we observe large improvements in calibration from using pre-training. In \Cref{fig:calibration} and \Cref{tab:calibresults}, we can see that RMS Calibration Error is at least halved on all datasets through the use of pre-training, with CIFAR-100 seeing the largest improvement. The same is true of the MAD error. In fact, the MAD error on CIFAR-100 is reduced by a factor of 4.1 with pre-training, which can be interpreted as the stated confidence being four times closer to the true frequency of occurrence.\looseness=-1

We find that these calibration gains are robust across pre-training datasets. With Places365 pre-training the RMS error is 3.1 on CIFAR-10, and with ImageNet pre-training the RMS error is 2.9; meanwhile, the baseline RMS error is 6.4. The gains are also complementary with the temperature tuning method of \citet{kilian}, which further reduces RMS Calibration Error from 4.15 to 3.55 for Tiny ImageNet when combined with pre-training. However, temperature tuning is computationally expensive and requires additional data, whereas pre-training does not require collecting extra data and can naturally and directly make the model more calibrated.

\section{Conclusion}

Although \citet{hepretrain} assert that pre-training does not improve performance on traditional tasks, for other tasks this is not so. On robustness and uncertainty tasks, pre-training results in models that surpass the previous state-of-the-art. For uncertainty tasks, we find pre-trained representations directly translate to improvements in predictive uncertainty estimates. \citet{hepretrain} argue that both pre-training and training from scratch result in models of similar accuracy, but we show this only holds for unperturbed data. In fact, pre-training with an untargeted adversary surpasses the long-standing state-of-the-art in adversarial accuracy by a significant margin. Robustness to label corruption is similarly improved by wide margins, such that pre-training alone outperforms certain task-specific methods, sometimes even after combining these methods with pre-training. This suggests future work on model robustness should evaluate proposed methods with pre-training in order to correctly gauge their utility, and some work could specialize pre-training for these downstream tasks. In sum, the benefits of pre-training extend beyond merely quick convergence, as previously thought, since pre-training can improve model robustness and uncertainty.\looseness=-1

\newpage

\subsubsection*{Acknowledgments}
We should like to thank David Wagner for his numerous suggestions. This research was partially supported by the Engineering Research Center Program through the National Research Foundation of Korea (NRF) funded by the Korean Government MSIT (NRF-2018R1A5A1059921).

\bibliography{biblio}
\bibliographystyle{icml2019}

\newpage

\begin{appendices}
\section{CIFAR-10-Related Classes Excluded from Downsampled ImageNet}
The ImageNet-1K classes that are related to CIFAR-10 are as follows:\\ n03345487, n03417042, n04461696, n04467665, n02965783, n02974003, n01514668, n01514859, n01518878, n01530575, n01531178, n01532829, n01534433, n01537544, n01558993, n01560419, n01580077, n01582220, n01592084, n01601694, n01608432, n01614925, n01616318, n01622779, n02123045, n02123159, n02123394, n02123597, n02124075, n02085620, n02085782, n02085936, n02086079, n02086240, n02086646, n02086910, n02087046, n02087394, n02088094, n02088238, n02088364, n02088466, n02088632, n02089078, n02089867, n02089973, n02090379, n02090622, n02090721, n02091032, n02091134, n02091244, n02091467, n02091635, n02091831, n02092002, n02092339, n02093256, n02093428, n02093647, n02093754, n02093859, n02093991, n02094114, n02094258, n02094433, n02095314, n02095570, n02095889, n02096051, n02096177, n02096294, n02096437, n02096585, n02097047, n02097130, n02097209, n02097298, n02097474, n02097658, n02098105, n02098286, n02098413, n02099267, n02099429, n02099601, n02099712, n02099849, n02100236, n02100583, n02100735, n02100877, n02101006, n02101388, n02101556, n02102040, n02102177, n02102318, n02102480, n02102973, n02104029, n02104365, n02105056, n02105162, n02105251, n02105412, n02105505, n02105641, n02105855, n02106030, n02106166, n02106382, n02106550, n02106662, n02107142, n02107312, n02107574, n02107683, n02107908, n02108000, n02108089, n02108422, n02108551, n02108915, n02109047, n02109525, n02109961, n02110063, n02110185, n02110341, n02110627, n02110806, n02110958, n02111129, n02111277, n02111500, n02111889, n02112018, n02112137, n02112350, n02112706, n02113023, n02113186, n02113624, n02113712, n02113799, n02113978, n01641577, n01644373, n01644900, n03538406, n03095699, n03947888.\\
We choose these WordNet IDs by using the WordNet hierarchy, though different class selections are conceivable.

\section{Evaluating Adversarial Robustness with Random Restarts}
PGD attacks with multiple random restarts and more iterations make for stronger adversaries. However, at the time of writing it is currently not standard to evaluate models with random restarts, even though restarts have been shown to reduce accuracy of adversarially trained models \cite{mosbach2018logit}. Other models completely break against adversaries using 100 steps. For completeness we include results with random restarts and more steps. An external evaluation by Marius Mosbach of our adversarially pre-trained CIFAR-10 model found that using 100 steps and 1,000 random restarts reduces accuracy to 52.9\%, using a 1,000 example subset of the CIFAR-10 test set. Compared to the baseline of normal training evaluated with a weaker adversary, adversarial pre-training remains 7.1\% higher in absolute accuracy. While it is not standard to evaluate with multiple random restarts, it is currently standard to evaluate with adversaries which take many steps. Adversaries with 100 steps and no restarts hardly affect the model's accuracy, in that accuracy changes from 57.4\% to 57.2\% or by only 0.2\%.

\section{Full Out-of-Distribution Detection Results}

We use the problem setup from \citet{hendrycks2019oe}. We use various datasets such as Gaussian Noise, Rademacher Noise, etc. Note that the ImageNet-21K dataset is the ImageNet-22K dataset with the ImageNet-1K classes held out. Results are in \Cref{tab:oodfull}.

 \begin{table*}[ht]
 % \small
  \caption{Out-of-distribution example detection for the maximum softmax probability baseline detector and the MSP detector after pre-training. All results are percentages and an average of 5 runs.}
 \centering
 \begin{tabularx}{\textwidth}{*{1}{>{\hsize=0.3\hsize}X} *{1}{>{\hsize=2.5cm}X }
 |*{2}{>{\hsize=0.6\hsize}Y}
 | *{2}{>{\hsize=0.6\hsize}Y} }
 % \toprule
 \multicolumn{2}{c}{}  & \multicolumn{2}{c}{AUROC $\uparrow$} &\multicolumn{2}{c}{AUPR  $\uparrow$}\\ \cline{3-6}
 $\mathcal{D}_\text{in}$ & \multicolumn{1}{l|}{$\mathcal{D}_\text{out}^\text{test}$} &
 {Normal} & {Pre-Training} & {Normal} & {Pre-Training} \\ \hline
 \parbox[t]{50mm}{\multirow{8}{*}{\rotatebox{90}{Tiny ImageNet}}}
 & Gaussian  & 49.4 & 67.4 & 15.2 & 21.1 \\
 & Rademacher& 70.7 & 75.0 & 23.0 & 25.5 \\
 & Blobs     & 76.2 & 69.5 & 28.2 & 23.1 \\
 & Textures  & 68.7 & 72.4 & 29.5 & 31.8 \\
 & SVHN      & 86.6 & 89.1 & 53.2 & 58.8 \\
 & Places365 & 76.8 & 74.6 & 36.8 & 31.8 \\
 & LSUN      & 73.2 & 71.6 & 30.4 & 27.4 \\
 & ImageNet-21K  & 72.7 & 71.7 & 29.9 & 28.5 \\
 \Xhline{0.5\arrayrulewidth} \multicolumn{2}{c|}{Mean} & {71.8} & {\textbf{73.9}} & {30.8} & {\textbf{31.0}} \\
 \Xhline{3\arrayrulewidth}
 \parbox[t]{50mm}{\multirow{8}{*}{\rotatebox{90}{CIFAR-10}}}
 & Gaussian  & 96.2 & 96.7 & 70.5 & 73.1 \\
 & Rademacher& 97.5 & 97.6 & 79.4 & 78.4 \\
 & Blobs     & 94.7 & 97.2 & 69.0 & 83.5 \\
 & Textures  & 88.3 & 93.7 & 56.6 & 70.4 \\
 & SVHN      & 91.8 & 95.7 & 63.7 & 76.9 \\
 & Places365 & 87.4 & 91.0 & 56.1 & 67.7 \\
 & LSUN      & 88.7 & 93.7 & 57.4 & 72.4 \\
 & CIFAR-100 & 87.1 & 90.7 & 54.1 & 65.4 \\
 \Xhline{0.5\arrayrulewidth} \multicolumn{2}{c|}{Mean} & {91.5} & {\textbf{94.5}} & {63.4} & {\textbf{73.5}} \\
 \Xhline{3\arrayrulewidth}
 \parbox[t]{50mm}{\multirow{8}{*}{\rotatebox{90}{CIFAR-100}}}
 & Gaussian  & 48.8 & 96.5 & 14.6 & 82.7 \\
 & Rademacher& 52.3 & 98.8 & 15.7 & 92.5 \\
 & Blobs     & 85.9 & 89.6 & 44.9 & 56.4 \\
 & Textures  & 73.5 & 79.7 & 33.1 & 44.1 \\
 & SVHN      & 74.5 & 79.6 & 32.0 & 48.5 \\
 & Places365 & 74.1 & 74.6 & 34.0 & 34.2 \\
 & LSUN      & 70.5 & 70.9 & 28.7 & 27.7 \\
 & CIFAR-10  & 75.5 & 75.3 & 34.5 & 35.8 \\
 \Xhline{0.5\arrayrulewidth} \multicolumn{2}{c|}{Mean} & {69.4} & {\textbf{83.1}} & {29.7} & {\textbf{52.7}} \\
 \Xhline{3\arrayrulewidth}
 \end{tabularx}

 \label{tab:oodfull}
 \end{table*}

\end{appendices}

% \section{Robustness to Online Decline}

\end{document}